\documentclass{article}

\usepackage[preprint]{corl_2023} 
\usepackage{graphicx}

\usepackage{svg}
\usepackage{paralist}

\title{GenORM: Generalizable One-shot Rope Manipulation with Parameter-Aware Policy}

\author{So Kuroki$^{1}$ Jiaxian Guo$^{1}$ Tatsuya Matsushima$^{1}$ Takuya Okubo$^{1}$ Masato Kobayashi$^{2}$\\
\textbf{Yuya Ikeda}$^{1}$ \textbf{Ryosuke Takanami}$^{1}$ \textbf{Paul Yoo}$^{1}$ \textbf{Yutaka Matsuo}$^{1}$ \textbf{Yusuke Iwasawa}$^{1}$\\
$^{1}$The University of Tokyo $^{2}$Osaka University
}

\begin{document}
\maketitle


\begin{abstract}
    Due to the inherent uncertainty in their deformability during motion, previous methods in rope manipulation often require hundreds of real-world demonstrations to train a manipulation policy for each rope, even for simple tasks such as rope goal reaching, which hinder their applications in our ever-changing world. 
    To address this issue, we introduce \emph{GenORM}, a framework that allows the manipulation policy to handle different deformable ropes with a single real-world demonstration. 
    To achieve this, we augment the policy by conditioning it on deformable rope parameters and training it with a diverse range of simulated deformable ropes so that the policy can adjust actions based on different rope parameters.
    At the time of inference, given a new rope, GenORM estimates the deformable rope parameters by minimizing the disparity between the grid density of point clouds of real-world demonstrations and simulations. With the help of a differentiable physics simulator, we require only a single real-world demonstration.
    Empirical validations on both simulated and real-world rope manipulation setups clearly show that our method can manipulate different ropes with a single demonstration and significantly outperforms the baseline in both environments (62\% improvement in in-domain ropes, and 15\% improvement in out-of-distribution ropes in simulation, 26\% improvement in real-world), demonstrating the effectiveness of our approach in one-shot rope manipulation. 
\end{abstract}
\vspace{-0.8em}

\vspace{-0.8em}
\section{Introduction}
\label{sec:introduction}
\vspace{-0.3cm}



Rope manipulation is performed in daily activities and is used in a wide range of tasks, including surgery~\cite{kitagawa2005effect, ganji2009robot}, knotting~\cite{suzuki2021air,schulman2016learning}, cable manipulation~\cite{jin2019robust, liu2023robotic}, and so on.
Despite the promising potential, learning a policy for rope manipulation is still challenging. It often necessitates numerous real-world demonstrations due to the complex dynamics of ropes, which encompass unpredictable deformability, elasticity, and friction during motion
~\cite{dom-survey2, dom-survey3}
. Nonetheless, the extensive volume of data required to train these models -- typically gathered through real robots -- can often be prohibitive~\cite{wang2014learning, nair2017combining}.

Recent research \cite{mehta2020active, rabinovitz2021unsupervised, leibovich2022validate, james2019sim,nachum2019multi} tried to alleviate this problem via Simulation-to-Real (Sim2Real) transfer, wherein policies are trained within simulation environments with unlimited data and subsequently applied to real-world tasks~\cite{matas2018sim}. Despite these efforts, mitigating Sim2Real gap continues to represent a protracted and complex problem within the robotics field~\cite{mehta2020active, rabinovitz2021unsupervised, leibovich2022validate, james2019sim,nachum2019multi,abeyruwan2023sim2real,pashevich2019learning,jakobi1995noise,mahler2017dex}. This challenge is exacerbated especially when dealing with deformable objects situated in unpredictable dynamic environments. A common approach to these complexities is to train a distinct policy for each rope, each possessing unique dynamics~\cite{lim2022real2sim2real, lee2022cavfm,angelina2019vpa}. This approach usually requires hundreds of real-world demonstrations, which are both time-consuming and costly to collect. This requirement for real-world data thus becomes a significant bottleneck for the scalability of robotics tasks, restricting the potential for broader application~\cite{lee2022cavfm,angelina2019vpa}.






In this paper, we propose a framework \emph{GenORM}, which allows the trained policy to manipulate different and even unseen ropes with a single real-world demonstration, \emph{i.e.}, one-shot generalization~\cite{rezende2016one}. In this way, we can reduce the real-world data requirements, thus reducing the deployment cost. Our key idea is to condition the policy with specific deformable rope parameters that can affect dynamics. Specifically, we select Young’s modulus and Poisson’s ratio~\cite{sanchez2018robotic, sengupta2020simultaneous, dom-pt1}
which are commonly utilized in modelling the deformation~\cite{sanchez2018robotic,sengupta2020simultaneous, dom-pt1}.
While these parameters hold significant potential in determining an object's deformability, their usage in policy training remains largely uncharted territory due to the inherent difficulties in estimating these values.
Our research represents an initial effort to integrate these parameters into manipulation policy training, thereby engendering a policy that is not only aware of rope-specific parameters but is also attuned to dynamics. As a result, the policy can adjust its action output for various ropes.
To train such an augmented manipulation policy, we randomly sample different rope parameters in the simulation and add them as the input of the policy network. This inclusion allows the policy to encompass a broader range of information on various dynamics parameters, utilizing a single network for deformable rope manipulation. 

During the deployment, we estimate rope parameters by leveraging Real-to-Simulation (Real2Sim) techniques \cite{sundaresan2022diffcloud, ma2022risp, le2023differentiable,huang2021plasticinelab, jatavallabhula2021gradsim, chendaxbench, heiden2021disect}. 
Concretely, with the capabilities of differentiable physics, we propose a gradient-based technique to estimate the rope parameters with only a single real-world demonstration. 
Our approach seeks to minimize the discrepancy of point clouds between the simulation and real-world environments. 
Contrary to prior Real2Sim methodologies \cite{sundaresan2022diffcloud, ma2022risp, le2023differentiable,huang2021plasticinelab, jatavallabhula2021gradsim, chendaxbench, heiden2021disect} that utilized estimated deformable rope parameters as simulation environmental parameters to learn the policy~\cite{lim2022real2sim2real}, our method uses these estimated parameters directly as the condition of the policy, enabling it to generalize to ropes with differing dynamics.


In summary, our contributions are as follows: (1) We propose a policy augmented with parameters that account for deformable rope characteristics (Young’s Modulus and Poisson’s ratio). We show that these parameters are effective enough to control the rope, and these can be estimated via designed gradient-based optimization given only a single real-world demonstration. (2) Empirical validation on both in-domain and out-of-distribution parameters of the dynamics demonstrate that GenORM excels in terms of its generalization capability across dynamics, even with only a single real-world demonstration. Specifically, GenORM provides 62\% improvement in in-domain ropes and 15\% improvement in out-of-distribution ropes in simulation, 26\% improvement in real-world, highlighting its superior adaptability and potential for broad applicability.

\section{Related Work}
\label{sec:related_work}

\vspace{-0.3cm}
\paragraph{Rope Manipulation}

Rope manipulation remains a practical yet challenging task, largely due to the uncertainty inherent in the deformability and friction of ropes in motion~\cite{sanchez2018robotic, dom-survey2, dom-survey3}.
Recent advancements in learning methods have begun addressing these tasks, showcasing substantial promise~\cite{lee2022cavfm, yan2020stn, dom-pt2, antonova2022rkhs}. 
Despite this progress, these methods are plagued with significant limitations.
The primary concerns are their substantial reliance on a large number of real-world demonstrations for effective learning and generalization capabilities for different scenarios~\cite{sanchez2018robotic, dom-survey2, dom-survey3, lee2022cavfm,angelina2019vpa}.
Various learning methods have been proposed that operate effectively on a limited amount of data~\cite{sundaresan2020learning, finn2017one}, although their performance was tends to be limited to specific objects or tasks.
In this study, our method offers extensive generalization capabilities and benefits from one-shot learning, significantly reducing the dependency on numerous real-world demonstrations. It also shows comparable performance to policies trained with over 100 real-wrold datasets.

\vspace{-0.3cm}
\paragraph{Sim2Real}
Numerous studies focus on filling the gap between simulation and real-world, domain randomization~\cite{mehta2020active, rabinovitz2021unsupervised, leibovich2022validate, james2019sim}, hierarchical learning~\cite{nachum2019multi, abeyruwan2023sim2real}, and data augmentation~\cite{pashevich2019learning}. In pursuit of dynamics generalization,~\cite{murooka2021exi} integrated both explicit and implicit dynamic parameters into their policy in a rigid-object push manipulation task. In the realm of deformable object manipulation,~\cite{kuroki2023collective, qi2022learning, lin2022planning} leverage differential physics to collect demonstrations on deformable object manipulation and subsequently train a policy through imitation learning to Sim2Real. Nonetheless, these models for deformable object manipulation have limitations of generalization ability for different objects due to it's difficulty in learning complex dynamics. To achieve this, we use parameters that account for deformable rope characteristics to train our policy.

\vspace{-0.3cm}
\paragraph{Real2Sim}
The demand for collecting realistic deformable object data in simulation scenarios is escalating. 
Much research is limited due to using specific devices or sensors~\cite{dom-pt1, dom-pt2}.
While learning-based methods to estimate parameters of deformable objects in simulation have been proposed~\cite{lim2022real2sim2real, yang2017learning}, these methods rely on huge datasets and are limited to specific objects. 
Thanks to the significant improvements in differentiable simulation~\cite{huang2021plasticinelab, jatavallabhula2021gradsim, chendaxbench, heiden2021disect}, several gradient-based methods have been introduced.
For example, \cite{sundaresan2022diffcloud} proposes differentiable point cloud sampling from mesh states.
\cite{ma2022risp} employed differentiable rendering methods. 
In this study, we aim to estimate deformable object parameters.
To ensure that the parameter captures the physical characteristics of objects, we select Young's modulus and Poisson's ratio to estimate, which is commonly used for modeling deformation as a linear and isotropic and homogeneous deformation model~\cite{sanchez2018robotic, sengupta2020simultaneous, dom-pt1}. 

\vspace{-0.3cm}
\paragraph{Real2Sim2Real}
Some research has been undertaken as Real2Sim2Real. \cite{wang2022real2sim2real} reconstructs high-quality meshes in simulation from real-world point clouds to achieve a robust policy in the real world.
\cite{wang2022real2sim2real} bridges the sim2real gap by optimizing parameters for tensegrity robots on a simulation platform using real data via gradient descent.
\cite{lim2022real2sim2real} estimates parameters with self-supervision and executes a sim2real transfer with a policy trained by the dataset collected with the estimated parameters. While the preceding research advocates for a Real2Sim2Real pipeline, they focus on dealing with specific dynamics, and few studies pursue the generalization ability for several dynamics.
\section{Proposed Method}
\label{sec:proposed_method}
\vspace{-0.3cm}
\begin{figure*}[t]
    \begin{center}
    \includegraphics[width=\textwidth]{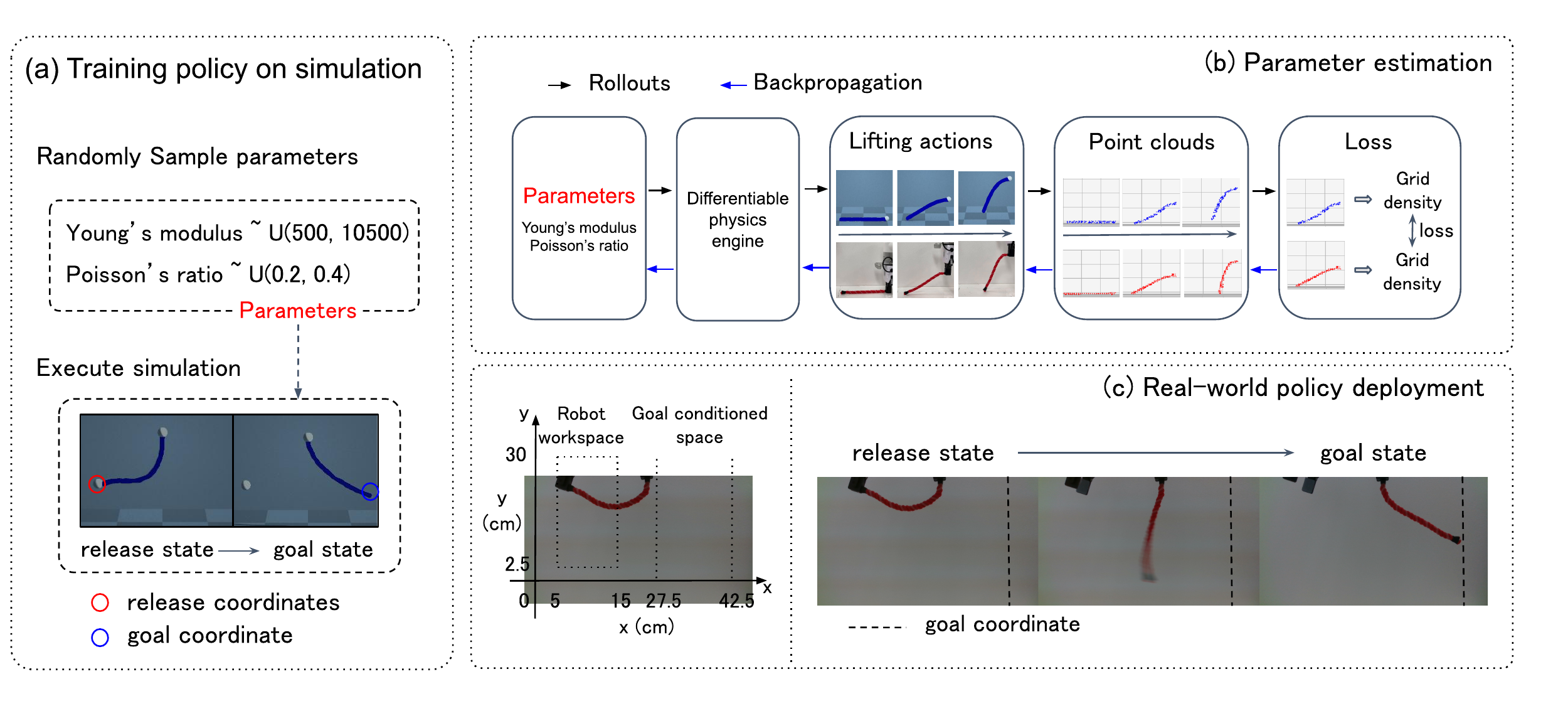}
    \vskip -0.2in
    \caption{Overview of our Pipeline: (a) We randomly sampled the parameters and set it to simulation where we collect release coordinates and goal coordinate. Receiving dataset, a simple MLP policy is trained. (b) For inference, deformable object parameters on simulation are estimated from real-world objects by taking loss from the grid density of point clouds between simulation and real-world via differentiable physics. (c) We transfer the policy onto real robots and conduct tests.}
    
    \label{fig:real2sim2real}
    \end{center}
    \vskip -0.30in
\end{figure*}

To achieve generalizable rope manipulation, we propose GenORM, and it mainly consists of two primal parts: (1) \emph{training augmented policy with diverse parameters on simulator} and (2) \emph{estimating rope parameters during the inference}. The entire GenORM pipeline is illustrated in Figure~\ref{fig:real2sim2real}. We will describe the problem setup with a canonical example and then present our framework in detail.

\subsection{{Canonical Task Description}}
\vspace{-0.6\baselineskip}
In this paper, we focus on the one-shot rope manipulation task which can be used for pick-and-place action for deformable object manipulation~\cite{deng2022deep, mo2023learning}, casting a rope~\cite{lim2022real2sim2real}, and motion control for brachiating robots~\cite{reda2022learning, davies2018tarzan}. Given a single demonstration of one rope and a goal coordinate $g$, a manipulation policy $\pi$ is trained to control a single robot arm to handle the end-point of this rope and guide it toward the releasing coordinates $r$ in mid-air. Upon release, the rope is expected to move under gravity towards a predetermined goal coordinate $g$.  As Figure~\ref{fig:real2sim2real} (c) shows, 
 the robot arm holds one end of the rope in the robot workspace ranging from 5cm to 15cm in the x-dimension and 2.5cm to 30cm in the y-dimension, and releases the rope to goal space ranging from 27.5cm to 42.5cm in the x-dimension. To choose the best releasing coordinates $r$, previous methods~\cite{deng2022deep, reda2022learning} usually take the goal coordinates $g$ as the input of the manipulation policy $\pi$ parameterized by $\theta$, and then output the releasing coordinates $r$, \emph{i.e.} $r = \pi_\theta(g)$. Owing to the intrinsic uncertainty of rope dynamics, these methods often necessitate hundreds of real-world demonstrations per-rope to adequately train such a policy. Next, we will present the details about how our method performs rope manipulation tasks given only a single demonstration.


\subsection{Training Policy with Diverse Parameters in Simulator} \label{sec:policy}
\vspace{-0.6\baselineskip}
The main challenge for learning a universally applicable rope manipulation policy arises due to the variation in physical mechanics across ropes. These attributional differences significantly influence the dynamics and thus the optimal release coordinates $r_k$, for a particular rope $k$, even when presented with identical goal coordinates $g_k$. To address this issue and learn such a generalizable rope manipulation policy,  our key idea is to augment the policy condition with deformable object parameters. This augmentation enables the policy to be aware of the dynamic variations and adjust its actions accordingly for ropes with different dynamics, thus enhancing its generalization ability for rope manipulation. Specifically, we propose to further condition policy on  Young’s modulus and Poisson’s ratio~\cite{sanchez2018robotic}. These two factors serve as critical determinants of a deformable object's behavior, representing stiffness and deformation characteristics, respectively. Both factors are frequently employed in the modeling of deformation \cite{sengupta2020simultaneous, dom-pt1}, and our paper is the first work to condition manipulation policy $\pi$ on these two factors to achieve generalizable rope manipulation.

As iteration cost in simulated environments is low, we opt to train our parameter-aware policy on simulation roll-outs first. Specifically,  we randomly sample release coordinate $r$, Young’s modulus parameter $p^y$, and Poisson’s ratio $p^l$ from predefined ranges, specifically [500, 10500] for Young’s modulus and [0.2, 0.4] for Poisson’s ratio.
During simulations runs, given the sampled rope parameters, we record the maximum reach distance of the bottom tip of the rope as the goal coordinate $g$. This data collection process is reiterated for $k$ steps to assemble the dataset $D = \{ p^y_{k}, p^l_{k}, r_{k}, g_{k} \}$. Taking  $g_k$, $p^y_{k}$, and $p^l_{k}$ as the input, the policy $\pi$ can estimate the corresponding releasing coordinate $\hat{r_k}$. Utilizing $g$ as hindsight, we train our policy $\pi$ by minimizing the following loss function:
\begin{equation}
\mathcal{L}_{\theta}\ =\  \frac{1}{N}\sum_{i=1}^{N} (  r_i - \pi_{\theta}(g_{i},  p^y_{i}, p^l_{i}))^2
\end{equation}
where $\theta$ is the trained parameter of the manipulation policy $\pi$, and $N$ is the batch size.
As we will show later, policy conditioned with deformable object parameters shows generalization not only to \emph{in-domain} parameters but also to \emph{out-of-distribution} parameters. We use simple Multi-layer Perceptrons (MLPs)~\cite{rumelhart1985learning} in our policy. We give more details about the policy in Appendix~\ref{appendix:params}.



\subsection{Gradient-based Real2Sim Parameter Optimization}
\vspace{-0.6\baselineskip}

After obtaining our augmented manipulation policy trained across an assortment of rope parameters, it remains challenging to apply this policy to real-world tasks due to the unavailability of specific rope parameters during real-world deployment. To mitigate this obstacle, we introduce the Real2Sim method, which estimates these parameters using a single real-world demonstration.
We begin by transforming a single real-world demonstration of a rope into point clouds, taking advantage of the inherent versatility of point clouds in representing diverse shapes and accommodating significant deformations, which makes them an ideal choice for simulating deformable objects. Utilizing differentiable physics~\cite{hu2019difftaichi,huang2021plasticinelab}, we adjust and estimate the rope parameters in the simulation to minimize the divergence between the grid density of point clouds from real-world demonstrations and simulations. As depicted in Figure~\ref{fig:real2sim2real} (b), our system is constructed on the framework of PlasticineLab, a differentiable simulator enabling gradient-based trajectory optimization through particle dynamics.
While performing predefined actions from identical initial states, we preserve point clouds in both the simulated and real-world environments. In these states, a robot lifts the rope vertically at a constant speed from its grasping point at the end of the rope on the desk. To capture these scenarios, we employ two RGB-D cameras in the real world and merge their respective point clouds.
Nonetheless, owing to the variances in the sequence and number of points between the simulator and real-world environments, a direct computation of the loss on point clouds is infeasible, hence, we cannot employ the prior knowledge within the differentiable physics simulation.

To bypass this hurdle, we represent the point clouds from both simulation and real-world demonstration into a grid density space (discretized into 64 $\times$ 64 $\times$ 64 in our paper) to synchronize them, regardless of their differing quantities and orders. Explicitly, each cloud's position is computed within the grid, assigning a weight to each grid cell based on its fractional position $f_x$ for superior density matching. Particularly, these weight assignments are attenuated based on cloud's distance to the cell's center, where the closer to the center, higher the weight, thereby promoting a smooth distribution of the particle's mass across neighbouring grid cells (More details are given in Appendix~\ref{appendix:weight}). This smooth distribution of the particle's mass enables successful alignment between the real-world and simulated scenarios, thereby utilizing the prior knowledge in the differentiable physics simulation to ensure an accurate loss that converges even with a single demonstration. By minimizing the L1 loss between the grid densities of the simulator and the real-world environments, with Young’s modulus and Poisson’s ratio serving as trainable variables, we can indirectly estimate these parameters through the differential simulator. We then condition these parameters directly on the pre-trained manipulation policy $\pi$ in Section \ref{sec:policy} to execute the rope manipulation task.
\section{Experiments}
\label{sec:experiments}

\vspace{-0.3cm}
In this section, we experimentally evaluate our method in both simulation and real world. Our experiments seek to address the following questions:
\begin{compactitem} 
    \item Can our gradient-based optimization using a designed loss function estimates the physical parameters of deformable objects (Young’s Modulus and Poisson’s ratio) given a limited demonstration? Does it better than a learning-based method? (\ref{subsec:sim2sim} and~\ref{subsec:real2sim}) 
    \item Does the estimated physical parameters help our parameter-aware policy to generalize diverse setups (in-domain, out-of-domain, and real-world deployment)? (\ref{subsec:ablation}, \ref{subsec:generalization_test}, and \ref{subsec:real_robot}) 
\end{compactitem}

\subsection{Setup}
\label{subsec:setup}
\vspace{-0.6\baselineskip}
\textbf{Rope:}
\begin{figure*}[t]
    \begin{center}
    \includegraphics[width=\textwidth]{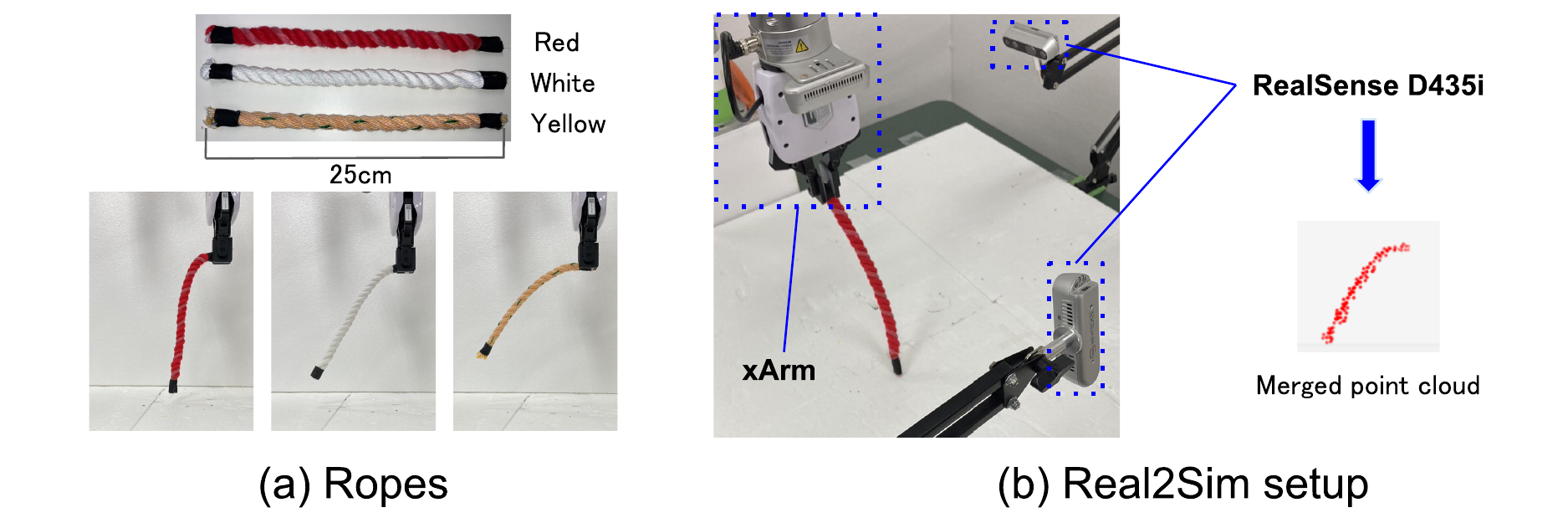}
    \vskip -0.2in
    \caption{(a) Three different type ropes; cotton rope (red), polyester rope (white), and polyethylene rope (yellow). The estimated Young’s modulus and Poisson’s ratio are [1779.38, 0.35], [3276.12, 0.346], and [8000.31, 0.36] for red, white, and yellow rope respectively. (b) Our experiments setup for Real2Sim: two Intel Realsense Depth Camera D435i and xArm 7.}
    \vskip -0.3in
    \label{fig:setup}
    \end{center}
\end{figure*}
We prepare three different types of ropes; cotton rope (red), polyester rope (white), and polyethylene rope (yellow). Figure~\ref{fig:setup} (a) visualizes them.


\textbf{Real2Sim:}
We estimate two parameters, Young's modulus and Poisson's ratio. Our hardware configuration, depicted in Figure~\ref{fig:setup} (b), incorporates two Intel Realsense Depth Camera D435i to obtain point clouds and xArm7 to manipulate rope. Two camera poses are calibrated to the xArm-world coordinate. We take a demonstration for each rope by executing the predefined lifting actions. Both real-world and simulation point clouds are transferred into (64, 64, 64) grid density. To train PointNet++ as a baseline, we randomly assign parameters to the simulation and execute the predefined actions to obtain the last state of the rope on the simulation. Using the 998 sets of parameters and the last state, PointNet++ is trained to output the parameter from the last state.

\textbf{Data Preparation and Training:} 
We train our policy on a synthesized dataset that covers diverse physical parameters of rope. Specifically, for simulation, we randomize Young’s modulus and Poisson’s ratio within the ranges [1500, 8200] and [0.34, 0.36], and for real2sim2real setup, we randomize them in [500, 10500] and [0.2, 0.4].
We also use two datasets for training Real2Sim2Real (R2S2R)~\cite{lim2022real2sim2real} baseline.  
(1) A dataset covering all three rope parameters, and (2) a dataset only covering the red rope parameter. 
We collect 2,271 demonstrations for simulation evaluation and 40,000 for realsim2real evaluation for all datasets. 
In addition, for the real2sim2real setup, we prepare a policy trained on only real-world dataset ($\pi_{RD}$): (1)  300 demonstrations for three ropes together and (2) 100 demonstrations with only the red rope's parameters. 
Initializing the policy weights with those trained in simulation, we fine-tuned our policy on 300 real-world demonstrations.

\subsection{Sim2Sim Performance}
\label{subsec:sim2sim}
\vspace{-0.6\baselineskip}
Table~\ref{tbl:sim2sim} compares our gradient-based and learning-based methods (Pointnet++) on the accuracy of sim2sim parameter estimation. 
We sampled five sets of ground-truth parameters and reported the mean and standard deviation. 
In addition to the estimation error of Young's modulus and Poisson's ratio, we calculate the Chamfer distance between point clouds of the last state with ground-truth parameters and estimated parameters.  
For the gradient-based method, we sample parameters as an initial state for each set of ground-truth parameters and proceed with optimization. 
PointNet++ estimates parameters from each last state after executing the predefined actions. 
We also test gradient-based optimization only with the last state loss for a fair comparison. 

 From the results, the gradient-based method yields a significantly smaller mean and standard deviation than PointNet++. 
 When comparing the gradient-based method, the gradient-based with the last step provide nearly identical performance with the gradient-based  with all steps.
 We utilize the gradient-based with all steps loss in subsequent experiments since it shows slightly more accuracy.

\begin{table}[!t]
\centering
\caption{Comparison of the accuracy for Sim2Sim parameter estimation. The mean and standard deviation of the difference between the estimated and ground-truth parameters and Chamfer distance between the last state with ground-truth parameters and estimated parameters.}
\begin{tabular}{cccc}
\hline
                             & \textbf{Young's modulus}  & \textbf{Poisson's ratio}      & \textbf{Chamfer}          \\
\textbf{Method}              & Mean of diff              & Mean of diff                  & \textbf{Distance}         \\ \hline
Gradient-based (all steps)   & $\mathbf{11.69 \pm 8.28}$ & $\mathbf{0.0032 \pm 0.0074}$  & $ \mathbf{0.0004   \pm 0.0005}$    \\ \hline
Gradient-based (last step)   & $13.74 \pm 6.75$          & $0.0067 \pm 0.0098$           & $0.0004   \pm 0.0004 $    \\ \hline
Pointnet++                   & $173.51\pm 200.14$        & $0.029  \pm 0.024 $           & $0.0025   \pm 0.0016 $    \\ \hline
\end{tabular}\label{tbl:sim2sim}
\vskip -0.15in
\end{table}

\subsection{Real2Sim Performance}
\label{subsec:real2sim}
\begin{figure*}[t]
    \begin{center}
    \includegraphics[width=\textwidth]{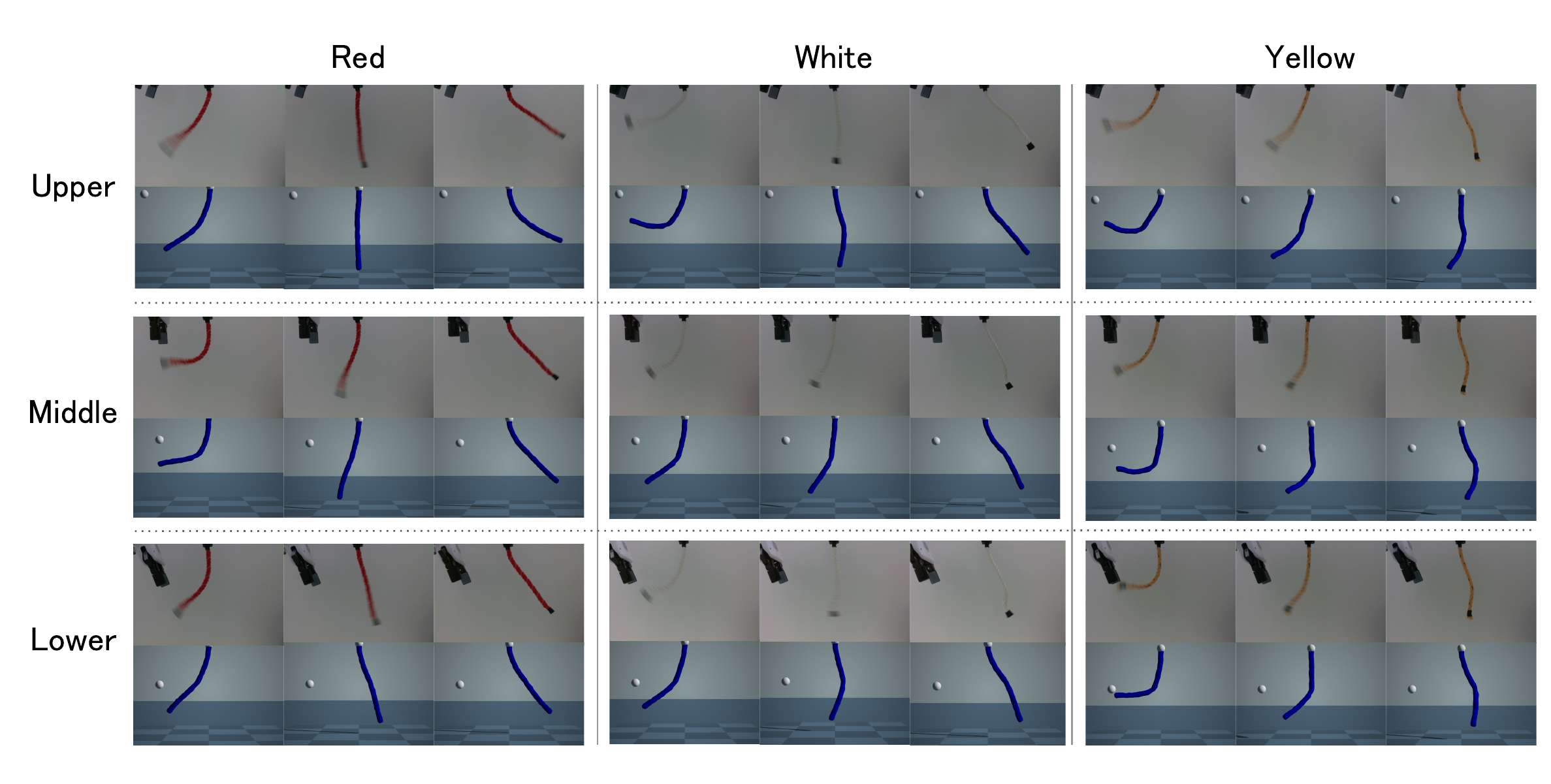}
    \vskip -0.2in
    \caption{As a qualitative assessment, we visualize rope trajectories in the real world and simulation with estimated parameters (Table~\ref{tbl:real2sim}) from the three different release points, upper, middle, and lower.}
    \vskip -0.2in
    \label{fig:compare}
    \end{center}
    
\end{figure*}
\vspace{-0.6\baselineskip}
We evaluate the parameter estimation of our method from a single real-world demonstration to a simulation. Table~\ref{tbl:real2sim} shows the estimated parameters and Chamfer distance between the last state of the real-world and the simulation with estimated parameters. The gradient-based method optimizes parameters from five randomly sampled parameters as the initial state for each rope. Pointnet++ uses point clouds of the last state to estimate parameters. For all ropes, Chamfer distance of the gradient-based method shows a smaller value than PointNet++.
The estimated Young’s modulus and Poisson’s ratio are [1779.38, 0.35], [3276.12, 0.346], and [8000.31, 0.36] for red, white, and yellow rope, respectively.
Although the Chamfer distance gap becomes smaller between our method and Pointnet++ for yellow rope, our method can directly estimate parameters without time-consuming data collection.
Figure~\ref{fig:compare} compare the trajectories between the real-world rope and simulation rope with estimated parameters from three levels of release coordinates, upper, middle, and lower. As a qualitative evaluation, two trajectories overlap except for yellow from the upper release. This may be because our method cannot express it correctly due to its strong elasticity.

\begin{table}[!t]
\centering
\caption{Comparison of the accuracy for Real2Sim parameter estimation. The mean and standard deviation of the estimated parameters and Chamfer distance between point clouds in real-worlds and point clouds with estimated parameters at the last state.}
\begin{tabular}{ccccc}
\hline
                 &                  & \textbf{Young's modulus}         & \textbf{Poisson's ratio}     & \textbf{Chamfer}               \\
\textbf{Rope}    &\textbf{Method}   & Estimated value                  & Estimated value              & \textbf{Distance}              \\ \hline
red              & Gradient-based   & $ \mathbf{1779.38 \pm 7.24}$     & $ \mathbf{0.35 \pm 0.038}$   & $\mathbf{0.034  \pm 0.0001}$   \\
                 & Pointnet++       & $509.83$                         & $0.30$                       & $0.092$                        \\ \hline
white            & Gradient-based   & $\mathbf{3276.12 \pm 3.71}$      & $ \mathbf{0.346 \pm 0.033}$  & $\mathbf{0.060  \pm 0.0001}$   \\
                 & Pointnet++       & $4660.12$                        & $0.30$                       & $0.067$                        \\ \hline
yellow           & Gradient-based   & $\mathbf{8000.31 \pm 48.04} $    & $ \mathbf{0.36 \pm 0.0036}$  & $\mathbf{0.053  \pm 0.0001}$   \\
                 & Pointnet++       & $7305.96$                        & $0.30$                       & $0.057$                        \\ \hline
\end{tabular}\label{tbl:real2sim}
\vskip -0.15in
\end{table}

\subsection{Ablation Study}
\label{subsec:ablation}
\vspace{-0.6\baselineskip}
To discern the contribution of each parameter to the task, we evaluate the accuracy by modulating inputs to the network. This assessment is conducted with four distinct input configurations: both Young's modulus and Poisson's ratio, only Young's modulus, only Poisson's ratio, and neither of these parameters. We randomly sample 30 new goal coordinates and parameters. Table~\ref{tbl:ablation} shows the best performance for the policy trained on both parameters, suggesting that the two parameters in conjunction are informative for the task. 

\begin{table}[!t]
\centering
\caption{Ablation study results. For training, we utilize four variations of inputs: both Young's modulus and Poisson's ratio, only Young's modulus, only Poisson's ratio, and neither of these parameters. We report the mean and standard deviation on 30 new goal coordinates. 
}
\begin{tabular}{cccc}
\hline
\textbf{Both parameters}      & \textbf{Young's mudulus}  & \textbf{Poisson's ratio}  & \textbf{No parameters}   \\ \hline
$ \mathbf{0.024 \pm 0.021}$   & $0.034 \pm 0.040$         & $0.062   \pm 0.058 $      & $0.062   \pm 0.059 $     \\ \hline
\end{tabular}\label{tbl:ablation}
\vskip -0.15in
\end{table}

\subsection{Generalization Test}
\label{subsec:generalization_test}
\vspace{-0.6\baselineskip}
To test generalizability to various ropes, we evaluate the policy on in-domain (ID) and out-of-distribution (OOD) parameters in simulation. For the ID test, we randomly sample 40 new goal coordinates with parameters of three ropes. For the OOD test, we randomly sample 40 new goal coordinates with parameters range [500, 1500] and [0.3, 0.33] for Young’s modulus and Poisson’s ratio, respectively. Table~\ref{tbl:id_ood} summarizes our findings. (1) Our method performs 62{\%} better than R2S2R (trained on three ropes) for ID. (2) Our method performs well even in the OOD setup. Specifically, we compare our method with R2S2R (trained on the red rope) for OOD. Even though the baseline is trained on the red rope's parameter, which is close to the OOD parameters' range, our method performs 15{\%} better than R2S2R (trained on the red rope). This suggests our parameter-conditioned policy has a stronger generalization ability against variable deformable object dynamics.

\begin{table}[!t]
\centering
\caption{Comparison of generalization for ID and OOD rope parameters. For inference, we randomly sample new 40 goal coordinates.}
\begin{tabular}{ccc}
\hline
\textbf{Method}                 & \textbf{ID}                   & \textbf{OOD}                     \\ \hline
Ours                            & $ \mathbf{0.031 \pm 0.050}$   & $ \mathbf{0.039 \pm 0.042}$      \\ \hline
R2S2R (Three ropes)             & $0.082 \pm 0.081$             & $0.096 \pm 0.104$                \\ \hline
R2S2R (Red rope)                & $0.047 \pm 0.033$             & $0.046 \pm 0.044$                \\ \hline
\end{tabular}\label{tbl:id_ood}
\vskip -0.15in
\end{table}

\subsection{Real Robot Experiments}
\label{subsec:real_robot}
\begin{figure*}[t]
    \begin{center}
    \includegraphics[width=\textwidth]{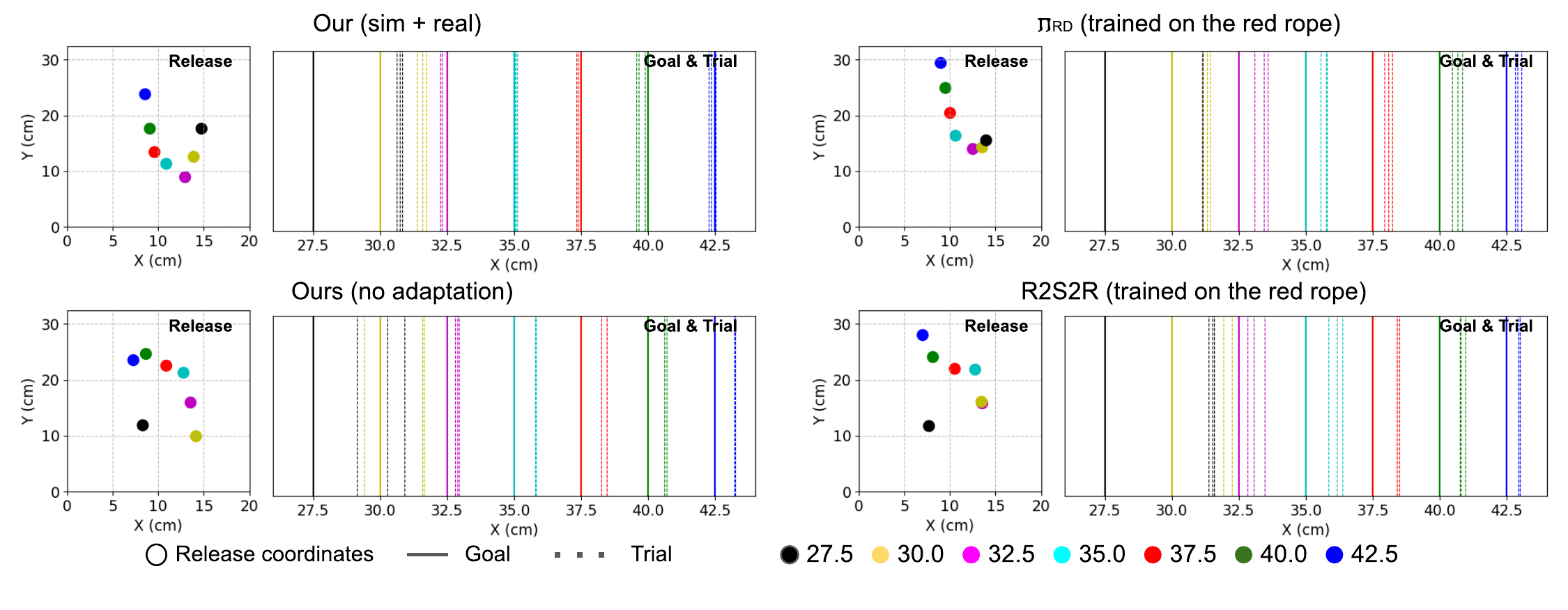}
    \vskip -0.2in
    \caption{The real-world deployment evaluation for red rope. For each policy, left figure proposes release coordinates (2-dimensions) and right figure illustrates the test performance for three trials. Each color is unified according to a goal coordination. The colors on the left and right sides of the figure are determined based on the goal coordination.}
     \vskip -0.1in
    \label{fig:realworld_evaluation}
    \end{center}
    
\end{figure*}
\vspace{-0.6\baselineskip}
Finally, we deploy our parameter-conditioned policy in real-world using the estimated parameters for each rope. We set a goal conditioned from [27.5, 30.0, 32.5, 35, 37.5, 40.0, 42.5] (cm) and tested the policy three times for each goal coordinate. We sumup the difference for each goal point and calculate its mean and standard deviation on three times trials. The results are presented in Table~\ref{tbl:sim2real}. For red rope, our method without adaptation perform 10{\%} better than $\pi_{RD}$ (trained on the red rope). This shows that our one-shot method, only using a single real-world demonstration, achieves almost the same performance with real-world policy trained with 100 demonstrations. With adaptation, our method (sim+real) performs 32{\%} better than $\pi_{RD}$ (trained on the red rope) and 44{\%} better than R2S2R (trained on the red rope). Moreover, our method (sim+real) outperforms baselines for all ropes performance. This is because our method can easily work in an in-domain range of parameters. For red rope comparison, Figure~\ref{fig:realworld_evaluation} illustrates the release coordinates (2-dimensions) and performance of three trials. Seeing the release coordinates of our (sim + real), we can confirm that our release coordinates without adaptation are adapted by training with the real-world dataset and shows the better smaller difference for each goal coordinate than without adaptation.


\begin{table}[!t]
\centering
\small
\setlength{\tabcolsep}{3pt}
\caption{Real-world deployment results. The goal coordinates are set from [27.5, 30.0, 32.5, 35, 37.5, 40.0, 42.5] (cm). We sum differences for each goal coordinate and report its mean and standard deviation on three times trials.}
\begin{tabular}{ccccccc}
\hline
                 & \textbf{Ours}                & \textbf{Ours}                 & \textbf{$\pi_{RD}$}    & \textbf{$\pi_{RD}$}       & \textbf{R2S2R}          & \textbf{R2S2R}      \\
\textbf{Rope}    & \textbf{(sim+real)}          & \textbf{(no adaptation)}      & \textbf{(Three ropes)} & \textbf{(Red rope)}       & \textbf{(Three ropes)}  & \textbf{(Red rope)} \\ \hline
red              & $ \mathbf{0.112 \pm 0.007}$  & $ \mathbf{0.148 \pm 0.012}$   & $0.426 \pm 0.017$      & $0.164 \pm 0.012$         & $0.318 \pm 0.004$       & $0.200 \pm 0.021$   \\ \hline
white            & $ \mathbf{0.076 \pm 0.025}$  & $ \mathbf{0.106 \pm 0.005}$   & $0.273 \pm 0.056$      & $0.512 \pm 0.039$         & $0.343 \pm 0.017$       & $0.464 \pm 0.038$   \\ \hline
yellow           & $ \mathbf{0.193 \pm 0.009}$  & $0.588 \pm 0.070$             & $0.297 \pm 0.113$      & $0.401 \pm 0.060$         & $0.364 \pm 0.061$       & $0.473 \pm 0.024$   \\ \hline
\end{tabular}\label{tbl:sim2real}
\vskip -0.15in
\end{table}
\section{Limitation and Conclusion}
\vspace{-0.3cm}
In this paper, we proposed a method for one-shot rope manipulation by combining pretrained parameter-conditioned policy with gradient-based parameter estimation method. We demonstrated that our method outperforms baseline both in-domain and out-of-distribution in simulation and real-world deployment. 
Our study has two key limitations: our methodology evaluation focuses only on rope manipulation and on simple tasks. Our Real2Sim approach, applicable to more flexible materials like dough, struggles to capture point cloud bias in real-world shape changes via RGB-D cameras. We plan to include a wider range of deformable objects and aim to test our method with complex tasks with dynamic deformation such as dough stretching or cloth hanging.


\clearpage

\bibliography{corl}

\clearpage
\appendix   
\section{Appendix}

\subsection{Summary of hyper-parameters}
\label{appendix:params}
We summarise hyper-parameters for differentiable physics Table~\ref{tab:hps1}, which we use for Sim2Sim and Real2sim, for PointNet++ Table~\ref{tab:hps2}, and for MLP Table~\ref{tab:hps1}

\begin{table}[h]
\centering
\setlength{\tabcolsep}{4pt}
\caption{Hyper-parameters for MLP and Differentiable physics.}
\begin{tabular}{lllll}
\multicolumn{2}{c}{MLP}             &  & \multicolumn{2}{c}{Differentiable physics}   \\ \cline{1-2} \cline{4-5} 
\textbf{Parameter} & \textbf{Value} &  & \textbf{Parameter} & \textbf{Value}          \\ \cline{1-2} \cline{4-5}
learning rate      & $1e-3$         &  & density loss       & $10$                    \\
batch size         & $256$          &  & learning rate      & $10$                    \\
clipnorm           & $0.1$          &  & optimizer          & Adam                    \\ \cline{1-2} \cline{4-5} 
\end{tabular}
\label{tab:hps1}
\vskip -0.1in
\end{table}

\begin{table}[h]
\centering
\setlength{\tabcolsep}{4pt}
\caption{Hyper-parameters for PointNet++}
\begin{tabular}{lllll}

\multicolumn{2}{c}{layer1}              &  & \multicolumn{2}{c}{layer2}                           \\ \cline{1-2} \cline{4-5}
\textbf{Parameter}  & \textbf{Value}    &  & \textbf{Parameter}   & \textbf{Value}                \\ \cline{1-2} \cline{4-5}
npoint              & $512$             &  & npoint               & $128$                         \\
radius              & $0.02$            &  & radius               & $0.02$                        \\
nsample             & $32$              &  & nsample              & $64$                          \\
mlp                 & [64, 64, 128]     &  & mlp                  & [128, 128, 256]               \\ \cline{1-2} \cline{4-5} 
\\

\multicolumn{2}{c}{layer3}              &  & \multicolumn{2}{c}{all}                             \\ \cline{1-2} \cline{4-5}
\textbf{Parameter}  & \textbf{Value}    &  & \textbf{Parameter}   & \textbf{Value}               \\ \cline{1-2} \cline{4-5}
npoint              &  -                &  & learning rate        & $1e-3$                       \\
radius              &  -                &  & batch size           & $16$                         \\
nsample             &  -                &  & clipnorm             & $0.1$                        \\
mlp                 & [256, 512, 1024]  &  & mlp                  & [256, 128]                   \\ \cline{1-2} \cline{4-5} 
\\
\end{tabular}
\label{tab:hps2}
\vskip -0.1in
\end{table}

\subsection{Weight for transferring point clouds to grid density}
\label{appendix:weight}
To assign a weight for each grid,  we calculate the grid according to the distance between the points and grid.  These weight assignments adhere to a bell-shaped curve, and the weight is defined as $0.75 - (f_x - 1) ^ 2$ for the closest grid cell, and either $0.5 * (1.5 - f_x) ^ 2$ or $0.5 * (f_x - 0.5) ^ 2$ is applied to the 26 grid cells surrounding the central cell.




\end{document}